\PassOptionsToPackage{unicode}{hyperref}
\PassOptionsToPackage{hyphens}{url}
\PassOptionsToPackage{dvipsnames,svgnames,x11names}{xcolor}
\documentclass[
]{article}
\usepackage{amsmath,amssymb}
\usepackage{lmodern}
\usepackage{arxiv}
\renewcommand{\shorttitle}{NICE Toolbox}
\usepackage{iftex}
\ifPDFTeX
  \usepackage[T1]{fontenc}
  \usepackage[utf8]{inputenc}
  \usepackage{textcomp} 
\else 
  \usepackage{unicode-math}
  \defaultfontfeatures{Scale=MatchLowercase}
  \defaultfontfeatures[\rmfamily]{Ligatures=TeX,Scale=1}
\fi
\IfFileExists{upquote.sty}{\usepackage{upquote}}{}
\IfFileExists{microtype.sty}{
  \usepackage[]{microtype}
  \UseMicrotypeSet[protrusion]{basicmath} 
}{}
\makeatletter
\@ifundefined{KOMAClassName}{
  \IfFileExists{parskip.sty}{%
    \usepackage{parskip}
  }{
    \setlength{\parindent}{0pt}
    \setlength{\parskip}{6pt plus 2pt minus 1pt}}
}{
  \KOMAoptions{parskip=half}}
\makeatother
\usepackage{xcolor}
\usepackage{graphicx}
\makeatletter
\def\maxwidth{\ifdim\Gin@nat@width>\linewidth\linewidth\else\Gin@nat@width\fi}
\def\maxheight{\ifdim\Gin@nat@height>\textheight\textheight\else\Gin@nat@height\fi}
\makeatother
\setkeys{Gin}{width=\maxwidth,height=\maxheight,keepaspectratio}
\makeatletter
\def\fps@figure{htbp}
\makeatother
\NewDocumentCommand\citeproctext{}{}
\NewDocumentCommand\citeproc{mm}{%
  \begingroup\def\citeproctext{#2}\cite{#1}\endgroup}
\makeatletter
 \let\@cite@ofmt\@firstofone
 \def\@biblabel#1{}
 \def\@cite#1#2{{#1\if@tempswa , #2\fi}}
\makeatother
\newlength{\cslhangindent}
\newlength{\csllabelwidth}
\newenvironment{CSLReferences}[2] 
 {\begin{list}{}{%
  \setlength{\itemindent}{0pt}
  \setlength{\leftmargin}{0pt}
  \setlength{\parsep}{0pt}
  \ifodd #1
   \setlength{\leftmargin}{\cslhangindent}
   \setlength{\itemindent}{-1\cslhangindent}
  \fi
  \setlength{\itemsep}{#2\baselineskip}}}
 {\end{list}}
\usepackage{calc}

\ifLuaTeX
\usepackage[bidi=basic]{babel}
\else
\usepackage[bidi=default]{babel}
\fi
\babelprovide[main,import]{american}

\def\languageshorthands#1{}
\ifLuaTeX
  \usepackage{selnolig}  
\fi
\IfFileExists{bookmark.sty}{\usepackage{bookmark}}{\usepackage{hyperref}}
\IfFileExists{xurl.sty}{\usepackage{xurl}}{} 
\hypersetup{
  pdftitle={NICE Toolbox: Nonverbal Interpersonal Communication
Exploration Toolbox},
  pdfauthor={Aleksandr Evgrashin, Carolin Schmitt, Timo Lübbing,
Ashutosh Jha, Buket Naz Zeren, Sophie K. Bauer, Johannes Kopf-Beck,
Anton K. G. Marx, Anamaria Semm, Gökce Ergün, Senya Polikovsky},
  pdflang={en-US},
  colorlinks=true,
  linkcolor={Maroon},
  filecolor={Maroon},
  citecolor={Blue},
  urlcolor={Blue},
  pdfcreator={LaTeX via pandoc}}

\title{NICE Toolbox: Nonverbal Interpersonal Communication Exploration
Toolbox}

\definecolor{c53baa1}{RGB}{83,186,161}
\definecolor{c202826}{RGB}{32,40,38}
\def \rorglobalscale {0.1}
\newcommand{\rorlogo}{%
\begin{tikzpicture}[y=1cm, x=1cm, yscale=\rorglobalscale,xscale=\rorglobalscale, every node/.append style={scale=\rorglobalscale}, inner sep=0pt, outer sep=0pt]
  \begin{scope}[even odd rule,line join=round,miter limit=2.0,shift={(-0.025, 0.0216)}]
    \path[fill=c53baa1,nonzero rule,line join=round,miter limit=2.0] (1.8164, 3.012) -- (1.4954, 2.5204) -- (1.1742, 3.012) -- (1.8164, 3.012) -- cycle;
    \path[fill=c53baa1,nonzero rule,line join=round,miter limit=2.0] (3.1594, 3.012) -- (2.8385, 2.5204) -- (2.5172, 3.012) -- (3.1594, 3.012) -- cycle;
    \path[fill=c53baa1,nonzero rule,line join=round,miter limit=2.0] (1.1742, 0.0669) -- (1.4954, 0.5588) -- (1.8164, 0.0669) -- (1.1742, 0.0669) -- cycle;
    \path[fill=c53baa1,nonzero rule,line join=round,miter limit=2.0] (2.5172, 0.0669) -- (2.8385, 0.5588) -- (3.1594, 0.0669) -- (2.5172, 0.0669) -- cycle;
    \path[fill=c202826,nonzero rule,line join=round,miter limit=2.0] (3.8505, 1.4364).. controls (3.9643, 1.4576) and (4.0508, 1.5081) .. (4.1098, 1.5878).. controls (4.169, 1.6674) and (4.1984, 1.7642) .. (4.1984, 1.8777).. controls (4.1984, 1.9719) and (4.182, 2.0503) .. (4.1495, 2.1132).. controls (4.1169, 2.1762) and (4.0727, 2.2262) .. (4.0174, 2.2635).. controls (3.9621, 2.3006) and (3.8976, 2.3273) .. (3.824, 2.3432).. controls (3.7505, 2.359) and (3.6727, 2.367) .. (3.5909, 2.367) -- (2.9676, 2.367) -- (2.9676, 1.8688).. controls (2.9625, 1.8833) and (2.9572, 1.8976) .. (2.9514, 1.9119).. controls (2.9083, 2.0164) and (2.848, 2.1056) .. (2.7705, 2.1791).. controls (2.6929, 2.2527) and (2.6014, 2.3093) .. (2.495, 2.3487).. controls (2.3889, 2.3881) and (2.2728, 2.408) .. (2.1468, 2.408).. controls (2.0209, 2.408) and (1.905, 2.3881) .. (1.7986, 2.3487).. controls (1.6925, 2.3093) and (1.6007, 2.2527) .. (1.5232, 2.1791).. controls (1.4539, 2.1132) and (1.3983, 2.0346) .. (1.3565, 1.9436).. controls (1.3504, 2.009) and (1.3351, 2.0656) .. (1.3105, 2.1132).. controls (1.2779, 2.1762) and (1.2338, 2.2262) .. (1.1785, 2.2635).. controls (1.1232, 2.3006) and (1.0586, 2.3273) .. (0.985, 2.3432).. controls (0.9115, 2.359) and (0.8337, 2.367) .. (0.7519, 2.367) -- (0.1289, 2.367) -- (0.1289, 0.7562) -- (0.4837, 0.7562) -- (0.4837, 1.4002) -- (0.6588, 1.4002) -- (0.9956, 0.7562) -- (1.4211, 0.7562) -- (1.0118, 1.4364).. controls (1.1255, 1.4576) and (1.2121, 1.5081) .. (1.2711, 1.5878).. controls (1.2737, 1.5915) and (1.2761, 1.5954) .. (1.2787, 1.5991).. controls (1.2782, 1.5867) and (1.2779, 1.5743) .. (1.2779, 1.5616).. controls (1.2779, 1.4327) and (1.2996, 1.3158) .. (1.3428, 1.2113).. controls (1.3859, 1.1068) and (1.4462, 1.0176) .. (1.5237, 0.944).. controls (1.601, 0.8705) and (1.6928, 0.8139) .. (1.7992, 0.7744).. controls (1.9053, 0.735) and (2.0214, 0.7152) .. (2.1474, 0.7152).. controls (2.2733, 0.7152) and (2.3892, 0.735) .. (2.4956, 0.7744).. controls (2.6016, 0.8139) and (2.6935, 0.8705) .. (2.771, 0.944).. controls (2.8482, 1.0176) and (2.9086, 1.1068) .. (2.952, 1.2113).. controls (2.9578, 1.2253) and (2.9631, 1.2398) .. (2.9681, 1.2544) -- (2.9681, 0.7562) -- (3.3229, 0.7562) -- (3.3229, 1.4002) -- (3.4981, 1.4002) -- (3.8349, 0.7562) -- (4.2603, 0.7562) -- (3.8505, 1.4364) -- cycle(0.9628, 1.7777).. controls (0.9438, 1.7534) and (0.92, 1.7357) .. (0.8911, 1.7243).. controls (0.8623, 1.7129) and (0.83, 1.706) .. (0.7945, 1.7039).. controls (0.7588, 1.7015) and (0.7252, 1.7005) .. (0.6932, 1.7005) -- (0.4839, 1.7005) -- (0.4839, 2.0667) -- (0.716, 2.0667).. controls (0.7477, 2.0667) and (0.7805, 2.0643) .. (0.8139, 2.0598).. controls (0.8472, 2.0553) and (0.8768, 2.0466) .. (0.9025, 2.0336).. controls (0.9282, 2.0206) and (0.9496, 2.0021) .. (0.9663, 1.9778).. controls (0.9829, 1.9534) and (0.9914, 1.9209) .. (0.9914, 1.8799).. controls (0.9914, 1.8362) and (0.9819, 1.8021) .. (0.9628, 1.7777) -- cycle(2.6125, 1.3533).. controls (2.5889, 1.2904) and (2.5553, 1.2359) .. (2.5112, 1.1896).. controls (2.4672, 1.1433) and (2.4146, 1.1073) .. (2.3529, 1.0814).. controls (2.2916, 1.0554) and (2.2228, 1.0427) .. (2.1471, 1.0427).. controls (2.0712, 1.0427) and (2.0026, 1.0557) .. (1.9412, 1.0814).. controls (1.8799, 1.107) and (1.8272, 1.1433) .. (1.783, 1.1896).. controls (1.7391, 1.2359) and (1.7052, 1.2904) .. (1.6817, 1.3533).. controls (1.6581, 1.4163) and (1.6465, 1.4856) .. (1.6465, 1.5616).. controls (1.6465, 1.6359) and (1.6581, 1.705) .. (1.6817, 1.7687).. controls (1.7052, 1.8325) and (1.7388, 1.8873) .. (1.783, 1.9336).. controls (1.8269, 1.9799) and (1.8796, 2.0159) .. (1.9412, 2.0418).. controls (2.0026, 2.0675) and (2.0712, 2.0804) .. (2.1471, 2.0804).. controls (2.223, 2.0804) and (2.2916, 2.0675) .. (2.3529, 2.0418).. controls (2.4143, 2.0161) and (2.467, 1.9799) .. (2.5112, 1.9336).. controls (2.5551, 1.8873) and (2.5889, 1.8322) .. (2.6125, 1.7687).. controls (2.636, 1.705) and (2.6477, 1.6359) .. (2.6477, 1.5616).. controls (2.6477, 1.4856) and (2.636, 1.4163) .. (2.6125, 1.3533) -- cycle(3.8015, 1.7777).. controls (3.7825, 1.7534) and (3.7587, 1.7357) .. (3.7298, 1.7243).. controls (3.701, 1.7129) and (3.6687, 1.706) .. (3.6333, 1.7039).. controls (3.5975, 1.7015) and (3.5639, 1.7005) .. (3.5319, 1.7005) -- (3.3226, 1.7005) -- (3.3226, 2.0667) -- (3.5547, 2.0667).. controls (3.5864, 2.0667) and (3.6192, 2.0643) .. (3.6526, 2.0598).. controls (3.6859, 2.0553) and (3.7155, 2.0466) .. (3.7412, 2.0336).. controls (3.7669, 2.0206) and (3.7883, 2.0021) .. (3.805, 1.9778).. controls (3.8216, 1.9534) and (3.8301, 1.9209) .. (3.8301, 1.8799).. controls (3.8301, 1.8362) and (3.8206, 1.8021) .. (3.8015, 1.7777) -- cycle;
  \end{scope}
\end{tikzpicture}
}

\usepackage[affil-it]{authblk}

\usepackage{orcidlink}
\author[1%
  ]{Aleksandr Evgrashin%
    }
\author[1%
  ]{Carolin Schmitt%
    }
\author[1%
  ]{Timo Lübbing%
    }
\author[1%
  ]{Ashutosh Jha%
    }
\author[1%
  ]{Buket Naz Zeren%
    }
\author[1,2%
  ]{Sophie K. Bauer%
    }
\author[1,2,3,4%
  ]{Johannes Kopf-Beck%
    }
\author[2,3%
  ]{Anton K. G. Marx%
    }
\author[2,3%
  ]{Anamaria Semm%
    }
\author[1%
  ]{Gökce Ergün%
    }
\author[1%
  ]{Senya Polikovsky%
    }

\affil[1]{Max Planck Institute for Intelligent Systems, Tübingen,
Germany%
    \,\protect\href{https://ror.org/04fq9j139}{\protect\rorlogo}\,%
  }
\affil[2]{Max Planck Institute of Psychiatry, Munich, Germany%
    \,\protect\href{https://ror.org/04dq56617}{\protect\rorlogo}\,%
  }
\affil[3]{LMU Munich, Department of Psychology, Germany%
    \,\protect\href{https://ror.org/05591te55}{\protect\rorlogo}\,%
  }
\affil[4]{German Center for Mental Health (DZPG), Partner Site Munich;
Augsburg, Germany%
    \,\protect\href{https://ror.org/00tkfw097}{\protect\rorlogo}\,%
  }
\date{10 July 2026}

\begin{document}
\maketitle

\begin{abstract}
Understanding human communication is a long-standing goal in
psychological research. People express themselves through many verbal
and nonverbal channels: posture, gesture, gaze, facial expression, and
speech. Measurement and annotation of these signals has traditionally
required time-consuming manual work by trained human labellers. The
\textbf{Nonverbal Interpersonal Communication Exploration (NICE)
Toolbox} is a free, open-source Python framework that automatically
measures these signals by combining existing state-of-the-art computer
vision and speech-processing methods into a single software package. It
includes support for multi-camera systems, interchangeable detection
algorithms, 2D and 3D visualization of detector outputs, evaluation on
the user's custom data, and export to third-party software. With the
release of this tool, we aim to make data exploration of interpersonal
communication signals scalable, reproducible, and accessible for
psychological research.
\end{abstract}

\section{Statement of need}\label{statement-of-need}

Labelling and annotation of nonverbal behaviour signals from video
recordings is one of the key methods of psychological research on
interpersonal communication
(\citeproc{ref-bakeman2011sequential}{Bakeman \& Quera, 2011}).
Traditionally, this is done manually by trained student assistants or
professional labellers rewatching videos multiple times. The process is
slow, difficult to scale, and results are hard to make consistent across
annotators (\citeproc{ref-vinciarelli2009social}{Vinciarelli et al.,
2009}). This also limits study dataset size, due to labelling budget and
time constraints.

Computer vision and speech-processing models exist for each of the
underlying signals (body pose, gaze, speech transcription, and others),
but they are distributed as separate research libraries with different
software dependencies, input/output formats, and configuration
conventions. Combining them into a working analysis workflow requires
machine-learning and software-engineering expertise, which usually lies
outside the scope of a psychology research project.

Furthermore, choosing among existing models to determine the most
suitable one for a specific study can be difficult for most
psychological researchers. Data recording setups vary widely in camera
placement, lighting, participant distance, appearance, interaction
style, and data formats. Algorithm rankings on public benchmarks often
do not transfer to a specific study, so researchers must compare
candidate models and hyperparameter configurations on their own data
before committing to one for their study
(\citeproc{ref-d2022underspecification}{D'Amour et al., 2022}). As of
yet, no available algorithm framework offers such an integrated
performance evaluation pipeline while allowing the combination of
different models to capture multimodal data streams.

\texttt{NICE\ Toolbox} is a Python framework for psychology researchers
that unifies data processing and model inference across a large set of
machine-learning algorithms. It uses a flexible configuration system for
defining custom user experiments, visualizations, evaluations, and
integration with third-party software. The measured signals are exported
as tabular data for further statistical analysis as part of a
psychological study.

\section{State of the field}\label{state-of-the-field}

Manual annotation tools such as \texttt{ANVIL}
(\citeproc{ref-kipp2001anvil}{Kipp, 2001}) and \texttt{ELAN}
(\citeproc{ref-wittenburg2006elan}{Wittenburg et al., 2006}) have long
been the standard for behavioural coding in psychological and linguistic
research. They provide advanced user interfaces for frame-accurate
labelling and are well integrated with the workflows of the communities
that use them, but the annotation itself remains a time-consuming and
labour-intensive process.

Automated open-source signal measurement tools usually cover only one
specific channel. \texttt{Motion\ Energy\ Analysis} (MEA)
(\citeproc{ref-ramseyer2020motion}{Ramseyer, 2020}) quantifies
frame-to-frame pixel change as a proxy for body movement and has been
widely used to study interpersonal synchrony in psychotherapy.
\texttt{OpenFace} (\citeproc{ref-Baltrusaitis2018}{Baltrusaitis et al.,
2018}) and \texttt{Py-Feat} (\citeproc{ref-cheong2023py}{Cheong et al.,
2023}) are open-source libraries for facial analysis used in
psychological research. \texttt{CrisperWhisper}
(\citeproc{ref-zusag2024}{Zusag et al., 2024}) and \texttt{WhisperX}
(\citeproc{ref-whisperx}{Bain et al., 2023}) are widely used for audio
transcription. Combining these tools into a multimodal analysis pipeline
is done by the researcher, who must adapt output formats, coordinate
systems, and sampling rates using custom post-processing scripts.

One effort to centralise these tools in a single hub is
\texttt{EnvisionBox} (\citeproc{ref-envisionbox}{EnvisionBox, 2026}). It
is a community-driven platform that curates open code modules,
tutorials, and datasets for multimodal analysis in social-interaction
research. Rather than providing a single unified software, it functions
as a shared storage for code examples demonstrating model inference and
data analysis.

Commercial platforms such as \texttt{FaceReader}
(\citeproc{ref-lewinski2014automated}{Lewinski et al., 2014}) and
\texttt{iMotions} (\citeproc{ref-imotions}{iMotions A/S, 2025}) provide
integrated multimodal analysis with polished user interfaces, but their
proprietary licensing, closed algorithms, and costs limit
reproducibility and accessibility for academic research. Their coverage
is also bounded by what the vendor supports: signals or models not
offered by the platform cannot be added by the user.

The closest comparison to \texttt{NICE\ Toolbox} is \texttt{DISCOVER}
(\citeproc{ref-hallmen2025discover}{Hallmen et al., 2025}), a
server-based back-end for computationally driven behaviour analysis.
Users drive it through its graphical front-end, the \texttt{NOVA}
annotation tool (\citeproc{ref-Heimerl2019}{Heimerl et al., 2019}), to
explore content, inspect recordings, and refine annotations.

Compared to \texttt{DISCOVER} and other alternatives,
\texttt{NICE\ Toolbox} focuses on automated large-scale dataset
processing without human intervention or refinement. Each user-created
configuration defines a computational pipeline in which detector outputs
from different modalities can feed into one another. The same pipeline
supports second-level measurements built on top of these outputs (for
example, kinematics from body pose, or interpersonal distance from
two-person tracking). Rather than committing to a single model,
\texttt{NICE\ Toolbox} provides an interchangeable collection of models
with a unified output structure. This allows users to evaluate
algorithms and choose the one that works best for their specific task.
\texttt{NICE\ Toolbox} supports multi-camera recordings and fuses
per-camera detections into a single 3D space through triangulation,
whereas most existing tools either accept only one camera per recording
or leave per-camera outputs unfused.

\section{Software design}\label{software-design}

\texttt{NICE\ Toolbox}'s design is based on three core principles: (1)
to provide a configuration-rich experiment design that lets
non-programmers run a full multimodal analysis without writing code, (2)
to maintain consistent data structures (components) across different
detectors, and (3) to reuse established open-source models and libraries
rather than reimplementing them (including MMPose
(\citeproc{ref-mmpose2020}{Contributors, 2020}), SAM 3D Body
(\citeproc{ref-sam3dbody}{Yang et al., 2026}), ETH-XGaze
(\citeproc{ref-Zhang2020ETHXGaze}{Zhang et al., 2020}), \texttt{Py-Feat}
(\citeproc{ref-cheong2023py}{Cheong et al., 2023}), and others).

\begin{figure}
\centering
\includegraphics[width=1\linewidth,height=\textheight,keepaspectratio]{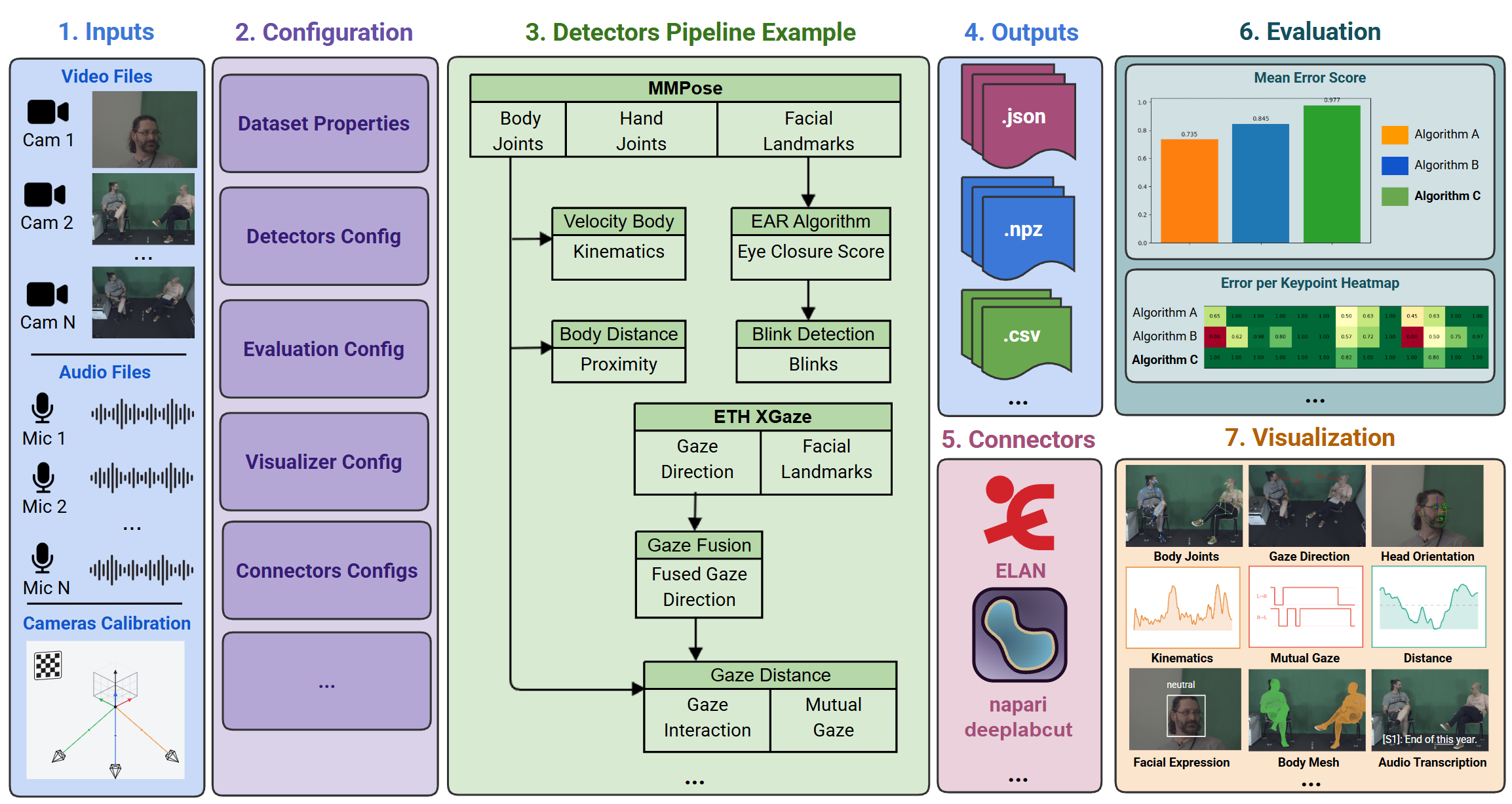}
\caption{\texttt{NICE\ Toolbox} experiment example. Multi-camera video
with optional audio and calibration files (1) are combined with
user-supplied TOML configuration (2) that describes the dataset,
requested detectors, evaluation, visualisation, and connectors. A
detector pipeline (3) connects primary models (e.g., MMPose for body
pose estimation, ETH-XGaze for gaze direction) with second-level
detectors (kinematics from body velocity, proximity from inter-body
distance, blinks from eye-closure ratios, and mutual gaze from fused
gaze directions across cameras). Results are saved per subject in
tabular formats (4), exported to external annotation tools such as
\texttt{ELAN} (5), and evaluated across candidate algorithms on the
user's own recordings (6), with metrics such as mean error score per
algorithm and per-keypoint error heatmap. Interactive 2D/3D
visualisation (7) shows final results for inspection and
diagnosis.\label{fig:software-design}}
\end{figure}

The toolbox is organised into modules (see the example experiment at
\autoref{fig:software-design}) with separate responsibilities:
\textbf{detectors} runs the computer vision and audio models on
recordings, \textbf{evaluation} computes configurable metrics based on
the user-specific dataset, \textbf{visualizer} provides the interactive
\texttt{rerun} (\citeproc{ref-RerunSDK}{Rerun Development Team, 2024})
3D visualizer, and \textbf{connectors} handles import and export to
third-party tools such as \texttt{ELAN} and \texttt{napari-deeplabcut}
(\citeproc{ref-napari_deeplabcut}{DeepLabCut, n.d.}) for ground-truth
annotation.

\begin{figure}
\centering
\includegraphics[width=1\linewidth,height=\textheight,keepaspectratio]{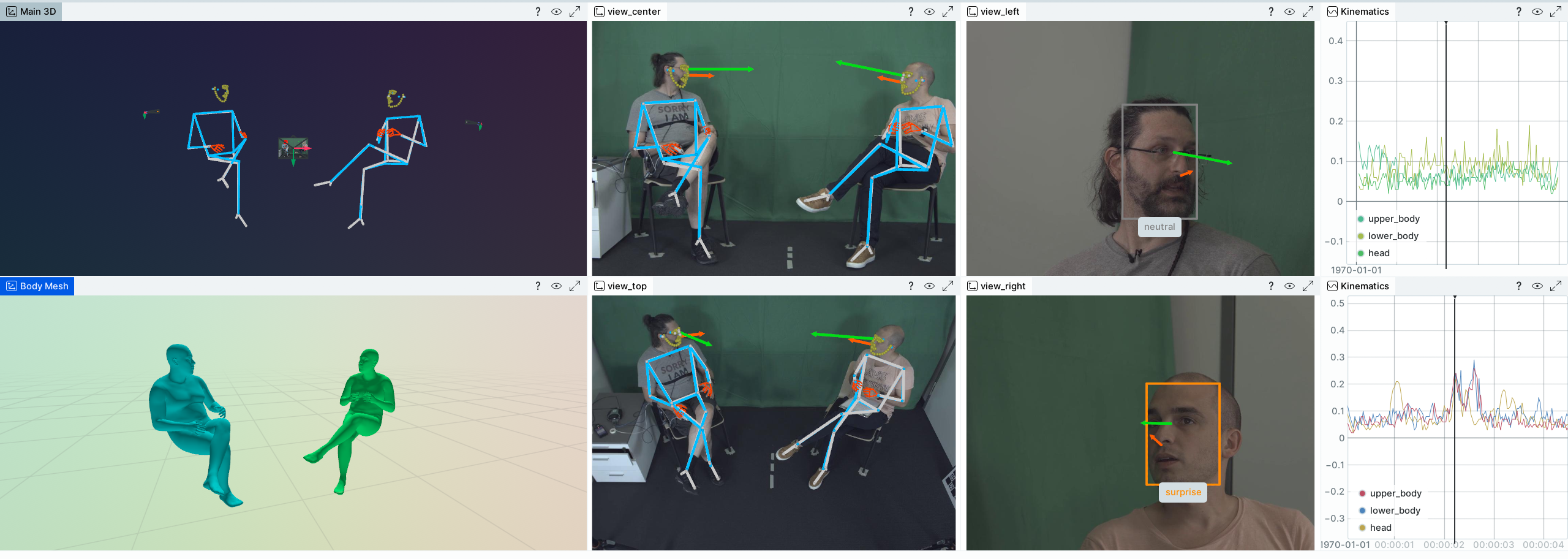}
\caption{An example layout of the interactive \texttt{NICE\ Toolbox}
rerun visualizer, showing detector results from a dyadic interaction:
triangulated body joints, body mesh, gaze direction, head orientation,
and synchronised kinematics time series. \label{fig:visualizer}}
\end{figure}

Users control the software through TOML configuration files that
describe algorithm parameters, evaluation metrics, visualizer layouts
(see \autoref{fig:visualizer}), and dataset structure. Configurations
are validated against schema definitions before execution, so errors are
caught before a long-running pipeline starts. To keep configuration
concise and flexible, \texttt{NICE\ Toolbox} supports variables and
named wildcards inside the configuration system. Detector algorithms
additionally support template inheritance, so users can define a
parameter set once and derive variants that override only the fields
that differ; for example, running the same pose estimator with different
confidence thresholds, or comparing two backbones that share the same
camera setup and I/O components.

Detectors can use the output of other detectors as their input, allowing
the construction of a directed computational graph. The order of
execution is defined by topological sorting of nodes. Nodes outputting
the same components are interchangeable and configurable by the user.
For example, a user can replace the MMPose model \texttt{body\_joints}
component with the SAM-3D-Body \texttt{body\_joints} component and use
it as input for the kinematics detector; the kinematics detector stays
the same, but its input can be easily replaced.

Most of the detectors run in their own isolated virtual or conda
environment. The main reason is that ML libraries often pin conflicting
versions of shared dependencies (e.g., \texttt{PyTorch}, \texttt{CUDA});
isolating each detector lets us use published models as-is without
patching them into a shared dependency tree. The main pipeline launches
each detector as a subprocess in its own environment, passes
configuration through a serialized file, and reads the detector's
outputs back from disk in the unified data format.

\section{Research impact statement}\label{research-impact-statement}

\texttt{NICE\ Toolbox} has been used to extract behavioural cues for
psychological research at the Max Planck Institute for Intelligent
Systems and by other collaborators. In work led by LMU Munich, the
toolbox was used to quantify movement synchrony from
pose-estimation-based kinematic displacement measures across 32 videos
of 16 parent--child dyads undergoing psychotherapy, distinguishing
genuine interactive coupling from coincidental covariation and comparing
free-play with structured-play conditions
(\citeproc{ref-Buaria:2026}{Buaria et al., 2026}). In a separate study
on emotional and behavioural responses to microaggressions in human--AI
interaction at work, the toolbox provided multimodal behavioural
measurements (facial expression, gaze, posture) across 482 interaction
sequences recorded from six synchronised cameras, collected alongside
self-reported responses (\citeproc{ref-Singh:2026}{Singh et al., 2026}).

\texttt{NICE\ Toolbox} is released under the GNU Affero General Public
License v3 (AGPL-3.0) and is publicly available on GitHub and Docker
Hub, with documentation, tutorials, and an example dataset.

\section*{AI usage disclosure}\label{ai-usage-disclosure}

Generative AI tools (Anthropic Claude Sonnet 4.6 and Opus 4.8) were used
during the development of this software and during manuscript editing.
All AI-generated code was reviewed by the original contributor and by a
second human reviewer in the merge-request process. All AI-assisted text
was reviewed and edited by the authors, who take full responsibility for
the final content.

\section*{Acknowledgements}\label{acknowledgements}

This work was partially supported by Max Planck \& Amazon Science Hub.

We thank the MPI-IS Software Workshop for their thoughtful feedback and
support during the project development.

\section*{References}\label{references}
\addcontentsline{toc}{section}{References}

\protect\phantomsection\label{refs}

\end{document}